\documentclass[runningheads,a4paper]{llncs}

\usepackage{amssymb}
\usepackage{graphicx}
\usepackage{epstopdf}
\usepackage{amsmath}
\usepackage{comment}
\usepackage{multirow}
\usepackage{arydshln}

\begin{document}

\mainmatter  

\title{Cross-modality image synthesis from unpaired data using CycleGAN}
\subtitle{Effects of gradient consistency loss and training data size}

\titlerunning{Cross-modality image synthesis from unpaired data using CycleGAN}

%

\author{Yuta Hiasa\inst{1}\and Yoshito Otake\inst{1}\and Masaki Takao\inst{2}\and \\Takumi Matsuoka\inst{1}\and Kazuma Takashima\inst{2}\and Aaron Carass\inst{3}\and \\ Jerry L. Prince\inst{3}\and Nobuhiko Sugano\inst{2}\and Yoshinobu Sato\inst{1}}

\institute{Graduate School of Science and Technology, Nara Institute of Science and Technology\\8916-5, Takayamacho, Ikomashi, Nara 630-0192, Japan \\\email{hiasa.yuta.ht7@is.naist.jp}
	\and Graduate School of Medicine, Osaka University
	\and Department of Electrical and Computer Engineering, Johns Hopkins University}

\authorrunning{Y. Hiasa et al.}


\maketitle

\begin{abstract}

CT is commonly used in orthopedic procedures. MRI is used along with CT to identify muscle structures and diagnose osteonecrosis due to its superior soft tissue contrast.  However, MRI has poor contrast for bone structures. Clearly, it would be helpful if a corresponding CT were available, as bone boundaries are more clearly seen and CT has a standardized (i.e., Hounsfield) unit. Therefore, we aim at MR-to-CT synthesis. While the CycleGAN was successfully applied to unpaired CT and MR images of the head, these images do not have as much variation of intensity pairs as do images in the pelvic region due to the presence of joints and muscles. In this paper, we extended the CycleGAN approach by adding the gradient consistency loss to improve the accuracy at the boundaries.  We conducted two experiments. To evaluate image synthesis, we investigated dependency of image synthesis accuracy on 1) the number of training data and 2) incorporation of the gradient consistency loss. To demonstrate the applicability of our method, we also investigated segmentation accuracy on synthesized images.

\end{abstract}

\section{Introduction}
Computed tomography (CT) is commonly used in orthopedic procedures. 
Magnetic resonance imaging (MRI) is used along with CT to identify muscle structures and diagnose osteonecrosis due to its superior soft tissue contrast \cite{cvitanic2004mri}. 
However, MRI has poor contrast for bone structures. 
It would be helpful if a corresponding CT were available, as bone boundaries are more clearly seen and CT has standardized (i.e., Hounsfield) units.
Considering radiation exposure in CT, it is preferable if we can delineate boundaries of both muscle and bones in MRI. 
Therefore, we aim at MR-to-CT synthesis. 

Image synthesis has been extensively studied using the patch-based learning \cite{torrado2016fast} as well as deep learning, specifically, convolutional neural networks (CNN) \cite{zhao2017whole} and generative adversarial networks (GAN) \cite{kamnitsas2017unsupervised}. The conventional approaches required the paired training data, i.e., images of the same patient from multiple modalities that are registered, which limited the application. A method recently proposed by Zhu et al. \cite{zhu2017unpaired}, called CycleGAN, utilizes the unpaired training data by appreciating the cycle consistency loss function. While CycleGAN has already applied to MR-to-CT synthesis \cite{wolterink2017deep}, all these previous approaches in medical image application targeted CT and MRI of the head in which the scan protocol (i.e., field-of-view (FOV) and the head orientation within the FOV) is relatively consistent resulting in a small variation in the two image distributions even without registration, thus a small number of training data set (20 to 30) allowed a reasonable accuracy. On the other hand, our target anatomy, the hip region, has larger variation in the anatomy as well as their pose (i.e., joint angle change and deformation of muscles).

Applications of image synthesis include segmentation.
Some previous studies aimed at segmentation of musculoskeletal structures in MRI \cite{gilles2010musculoskeletal,ranzini2017joint}, but 
the issues in these studies were the requirement for multiple sequences and devices.
Another challenge in segmentation of MRI is that there is no standard unit as in CT. Therefore, manually traced label data are necessary for training of each sequence and each imaging device. Thus, MR-to-CT synthesis realizes modality independent segmentation \cite{hamarneh2008simulation}.

In this study, we extend the CycleGAN approach by adding the gradient consistency (GC) loss to encourage edge alignment between images in the two domains and using an order-of-magnitude larger training data set (302 MR and 613 CT volumes) in order to overcome the larger variation and improve the accuracy at the boundaries.
We investigated dependency of image synthesis accuracy on 1) the number of training data and 2) incorporation of the GC loss. To demonstrate the applicability of our method, we also investigated a segmentation accuracy on synthesized images.

\section{Method}

\subsection{Materials}

\begin{figure}[!bt]
	\centering
	\includegraphics[width=0.95\textwidth]{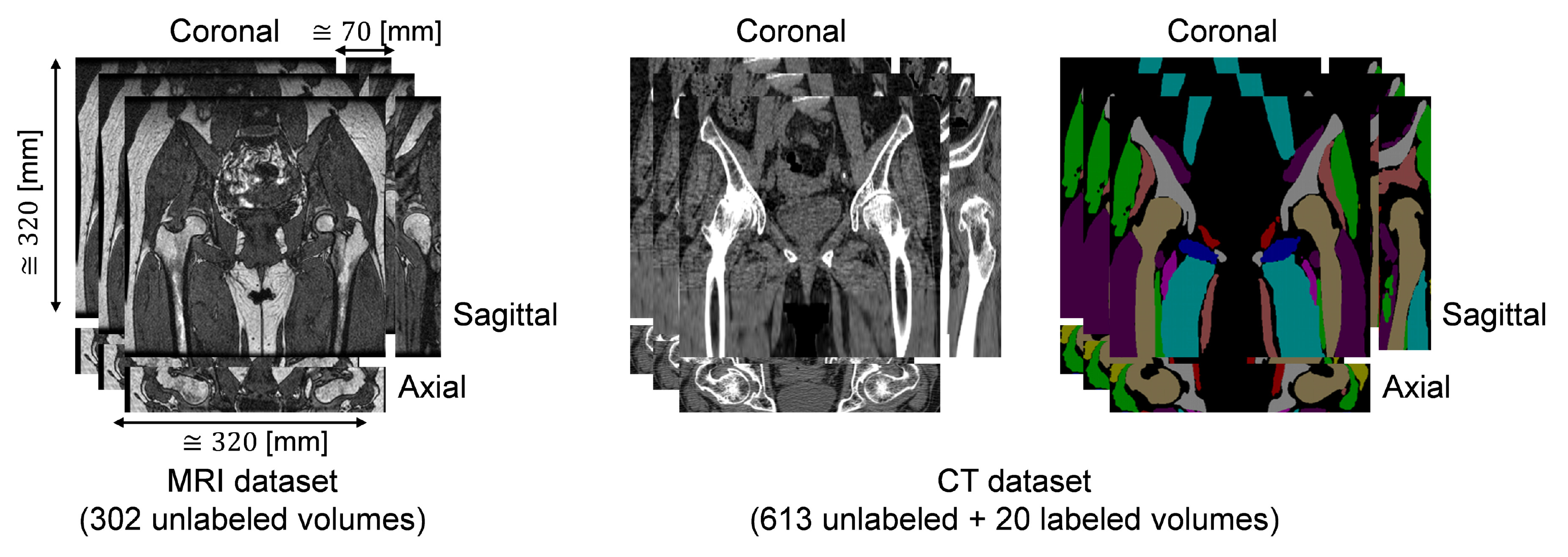}
	\caption{Training datasets used in this study. MRI dataset consists of 302 unlabeled volumes and CT dataset consists of 613 unlabeled and 20 labeled volumes. N4ITK intensity inhomogeneity correction \cite{tustison2010n4itk} was applied to all MRI volumes. Two datasets have similar field-of-view, although these are not registered.
	}
	\label{fig:dataset}
\end{figure}

The datasets we used in this study are MRI dataset consisting of 302 unlabeled volumes and CT dataset consisting of 613 unlabeled, and 20 labeled volumes which are associated with manual segmentation labels of 19 muscles around hip and thigh, pelvis, femur and sacrum bones. Patients with metallic artifact due to implant in the volume were excluded. As an evaluation dataset, we also used other three sets of paired MR and CT
volumes, and 10 MR volumes associated with manual segmentation labels of gluteus medius and minimus muscles, pelvis and femur bones, as a ground truth.
MR volumes were scanned in the coronal plane for diagnosis of osteonecrosis by a 1.0T MR imaging system. 
The T1-weighted volumes were obtained by 3D spoiled gradient recalled echo sequence (SPGR)
with a repetition time (TR) of 7.9 ms, echo time (TE) of 3.08 ms, and flip angle of 30. The field of view was 320 mm, and the matrix size was 256$\times$256. The slab thickness was 76 mm, and the slice thickness was 2 mm without an inter-slice gap.
CT volumes were scanned in the axial plane for diagnosis of the patients subjected to total hip arthroplasty (THA) surgery. The field of view was 360$\times$360 mm and the matrix size was
512$\times$512. The slice thickness was 2.0 mm for the
region including pelvis and proximal femur, 6.0 mm
for the femoral shaft region, and 1.0 mm for the distal
femur region. 
In this study, the CT volumes were cropped and resliced so that the FOV resembles that of MRI volumes, as shown in Figure \ref{fig:dataset}, and then resized to 256$\times$256.

\subsection{Image synthesis using CycleGAN with gradient-consistency loss}

\begin{figure}[!bt]
	\centering
	\includegraphics[width=0.60\textwidth]{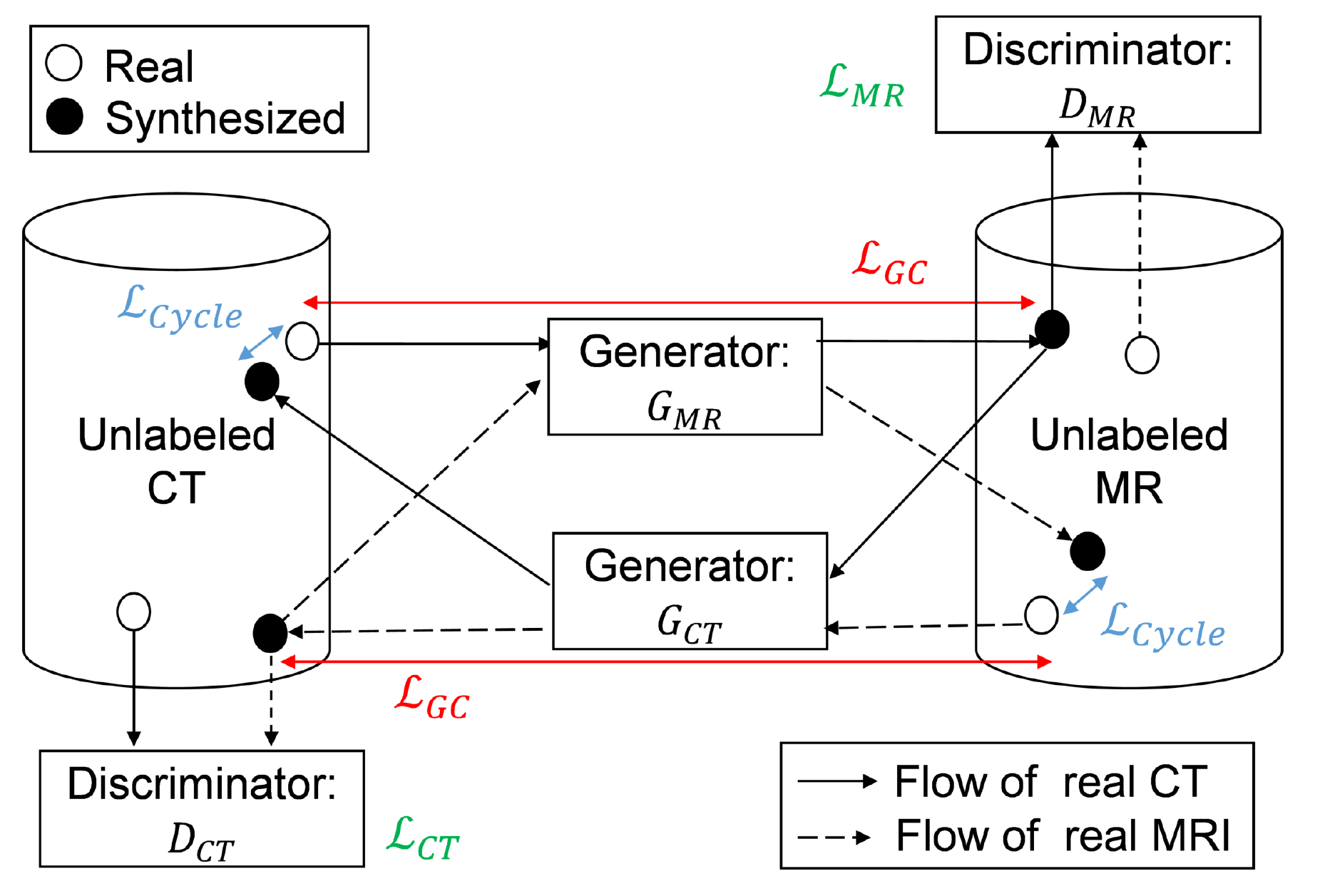}
	\caption{Workflow of the proposed method. $G_{CT}$ and $G_{MR}$ are generator networks that translate MR to CT images, and CT to MR images, respectively. $D_{CT}$ and $D_{MR}$ are discriminator networks to distinguish between real and synthesized images. The cycle consistency loss $\mathcal{L}_{Cycle}$ is a regularization term defined by the difference between real and reconstructed image. To improve the accuracy at the edges, loss function is regularized by gradient consistency loss $\mathcal{L}_{GC}$. }
	\label{fig:overview}
\end{figure}

The underlying algorithm of the proposed MR-to-CT synthesis follows that of Zhu et al \cite{zhu2017unpaired} which allows to translate an image from CT domain to MR domain without pairwise aligned CT and MR training images of the same patient. The workflow of the proposed method is shown in Figure \ref{fig:overview}. 
The networks $G_{CT}$ and $G_{MR}$ are generators to translate real MR and CT images to synthesized CT and MR images, respectivery. The networks $D_{CT}$ and $D_{MR}$ are discriminators to distinguish between real and synthesized images. 
While discriminators try to distinguish synthesized images by maximizing adversarial losses $\mathcal{L} _{CT}$ and $\mathcal{L} _{MR}$, defined as
\begin{eqnarray}
\mathcal{L} _{CT} &=& \textstyle \sum_{x\in I_{CT}} \log D_{CT}(x) + \sum_{y\in I_{MR}} \log ( 1-D_{CT}(G_{CT}(y))), \\
\mathcal{L}_{MR} &=& \textstyle \sum_{y\in I_{MR}} \log D_{MR}(y) + \sum_{x\in I_{CT}} \log (1-D_{MR}(G_{MR}(x))),
\end{eqnarray}
generators try to synthesize images which is indistinguishable from the target domain by minimizing these losses. Where $x$ and $y$ are images from domains $I_{CT}$ and $I_{MR}$.
However, networks with large capacity have potential to converge to the one that translate the same set of images from source domain to any random permutation of images in the target domain. Thus, adversarial losses alone cannot guarantee that the learned generator can translate an individual input to a desired corresponding output. Therefore, the loss function is regularized by cycle
consistency, which is defined by the difference between real and reconstructed image, which is the inverse mapping of the synthesized image \cite{zhu2017unpaired}. The cycle consistency loss $\mathcal{L}_{Cycle}$ is defined as
\begin{eqnarray}
\mathcal{L}_{Cycle} &=& \textstyle \sum_{x\in I_{CT}} |G_{CT}(G_{MR}(x)) - x| + \sum_{y\in I_{MR}} |G_{MR}(G_{CT}(y)) - y|
\end{eqnarray}

We extended the CycleGAN approach by explicitly adding the gradient consistency loss between real and synthesized images to improve the accuracy at the boundaries. 
The gradient correlation (GC) \cite{penney1998comparison} has been used
as a similarity metric in the medical image registration, which is defined by the normalized cross correlation between two images. Given gradients in horizontal and vertical directions of thes two images, $A$ and $B$, GC is defined as
\begin{eqnarray}
GC(A, B) &=& \frac{1}{2}\{NCC(\nabla_x A, \nabla_x B) + NCC(\nabla_y A, \nabla_y B) \} 
\label{eq:gc} \\ 
\mathrm{where},\  NCC(A, B) &=& \frac{ \sum_{(i,j)}^{} (A-\bar{A}) (B-\bar{B})}{ \sqrt{ \sum_{(i,j)}^{} (A-\bar{A})^2 } \sqrt{ \sum_{(i,j)}^{} (B-\bar{B})^2 }} \nonumber
\end{eqnarray}
and $\nabla_x$ and $\nabla_y$ are the gradient operator of each direction, $\bar{A}$ is the mean value of $A$. 
We formulate the gradient-consistency loss $\mathcal{L}_{GC}$ as  
\begin{eqnarray}
\mathcal{L}_{GC} &=& \frac{1}{2}\{ \sum_{x\in I_{CT}} (1-GC(x, G_{MR}(x)))+ \sum_{y\in I_{MR}} (1-GC(y, G_{CT}(y)))\} 
\end{eqnarray}
Finally, our objective function is defined as:
\begin{eqnarray}
\mathcal{L}_{total} &=& \mathcal{L}_{CT} + \mathcal{L}_{MR} + \lambda_{Cycle} \mathcal{L}_{Cycle} + \lambda_{GC} \mathcal{L}_{GC}
\end{eqnarray}
where $\lambda_{Cycle}$ and  $\lambda_{GC}$ are weights to balance each loss. Then, we solve:
\begin{eqnarray}
\hat{G}_{MR}, \hat{G}_{CT} = \arg \min_{G_{CT},G_{MR}} \max_{D_{CT},D_{MR}} \mathcal{L}_{total} 
\end{eqnarray}

In this paper, we used 2D CNN with 9 residual blocks for generator, similar to the one proposed in \cite{johnson2016perceptual}.
For discriminators, we used $70 \times 70$ PatchGAN \cite{isola2017image}. We replaced the Eq. (1) and Eq. (2) by least-squares loss as in \cite{mao2016multi}. These settings follows \cite{zhu2017unpaired,wolterink2017deep}.
The CycleGAN was trained using Adam \cite{kingma2014adam} for the first $1\times10^5$ iterations at fixed learning rate of 0.0002, and the last $1\times10^5$ iterations at learing rate which linearly reducing to zero. The balancing weights were empirically determined as $\lambda_{Cycle} = 3$ and $\lambda_{GC} = 0.3$. CT and MR volumes are normalized such that intensity of [-150, 350] HU and [0, 100] are mapped to [0, 255], respectively.

\section{Result}
\subsection{Quantitative evaluation on image synthesis}

To evaluate image synthesis, we investigated dependency of the accuracy on the number of training data and with or without the GC loss. 
The CycleGAN was trained with datasets of different sizes, i) 20 MR and 20 CT volumes, ii) 302 MR and 613 CT volumes, and both with and without GC loss. 
We conducted two experiments. The first experiment used three sets of paired MR and CT volumes of the same patient for test data. Because availability of paired MR and CT volumes was limited, we conducted the second experiment in which unpaired 10 MR and 20 CT volumes were used.

\begin{figure}[!bt]
	\centering
	\includegraphics[width=0.9\textwidth]{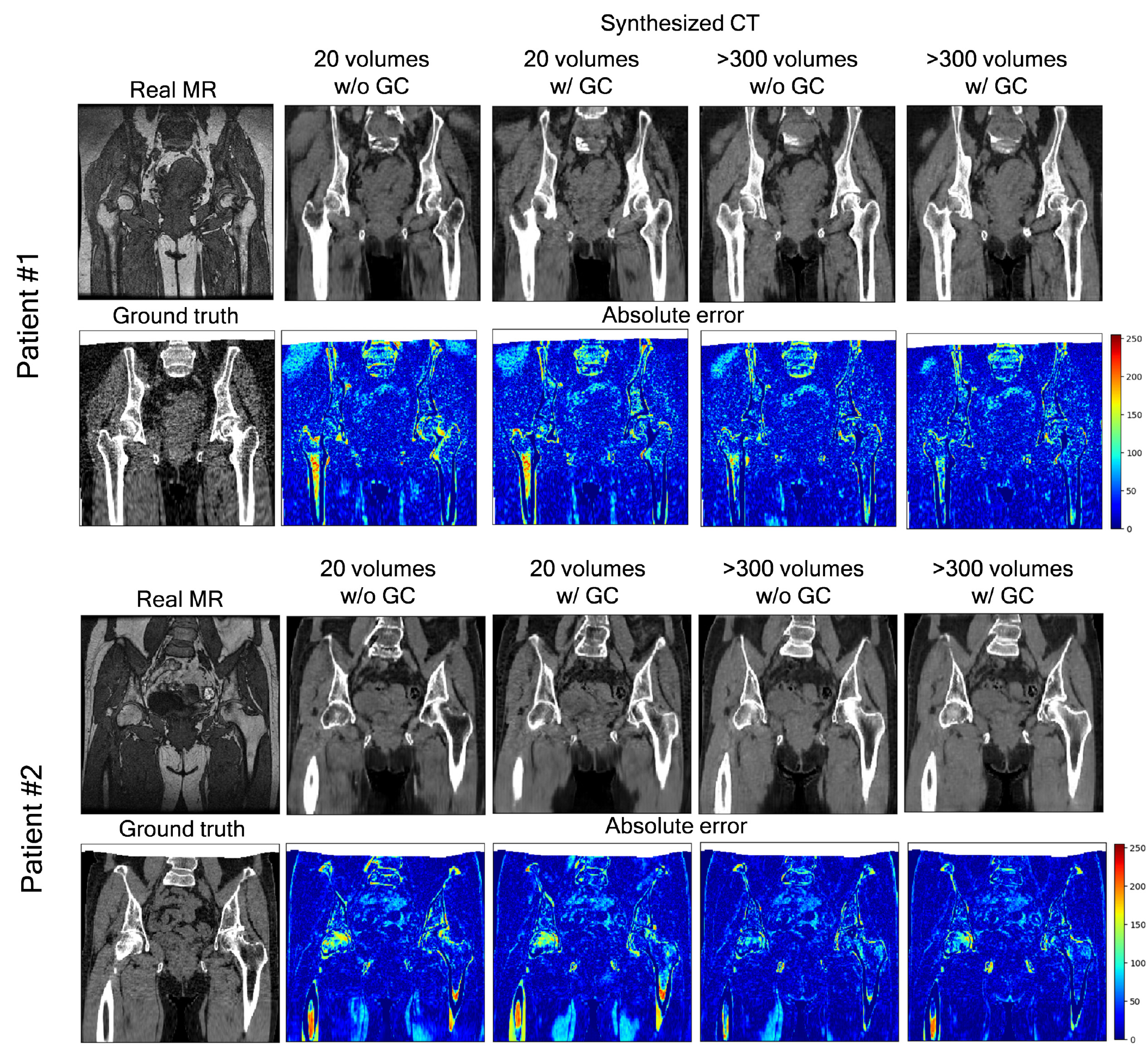}
	\caption{Representative results of the absolute error between the ground truth paired CT and synthesized CT from two patients. Since the FOV of MR and CT volumes are slightly different, there is no corresponding region near the top edge of the ground truth volumes (filled with white color). This area was not used for evaluation.}
	\label{fig:vis_paired}
\end{figure}

In the first experiment, we evaluated synthesized CT by means of mean absolute error (MAE) and peak-signal-to-noise ratio (PSNR) [dB] between synthesized CT and ground truth CT, both of which were normalized as mentioned in 2.2. The ground truth CT here is a CT registered to the MR of the same patient.
CT and MR volumes were aligned using landmark-based registration as initialization, and then aligned using rigid and non-rigid registration. The results of MAE and PSNR are shown in Table \ref{tab:mae}. PSNR is calculated as $PSNR = 20 \log_{10}\frac{255}{\sqrt{MSE}}$, where MSE is mean squared error. 
The average of MAE decreased and PSNR increased according to the increase of training data size and inclusion of GC loss, respectively. Fig \ref{fig:vis_paired} shows representative results.

\begin{table}[t]
	\centering
	\caption{Mean absolute error (MAE) and Peak-signal-to-noise ratio (PSNR) between synthesized and real CT volumes.}
	\label{tab:mae}
	\begin{tabular}{l|l|l|l|l|l|}
		& & \multicolumn{2}{|c|}{20 volumes} & \multicolumn{2}{|c|}{$>$300 volumes} \\ 
		& & \multicolumn{1}{|c|}{w/o GC} & \multicolumn{1}{|c|}{/w GC} & \multicolumn{1}{|c|}{w/o GC} & \multicolumn{1}{|c|}{/w GC} \\ \hline
		\multirow{4}{*}{MAE}& Patient \#1 & 30.121 & 30.276 & 26.899 & 26.388 \\
		& Patient \#2 & 26.927 & 26.911 & 22.319 & 21.593 \\
		& Patient \#3 & 33.651 & 32.155 & 29.630 & 28.643 \\ \cdashline{2-6}
		& Average $\pm$ SD     & 30.233 $\pm$ 2.177 & 29.781 $\pm$ 1.777 & 26.283 $\pm$ 1.367 & 25.541 $\pm$ 1.129 \\ \hline

		\multirow{4}{*}{PSNR} & Patient \#1 & 14.797 & 14.742 & 15.643 & 15.848  \\
		& Patient \#2 & 15.734 & 15.628 & 17.255 & 17.598 \\
		&Patient \#3 & 14.510 & 14.820 & 15.674 & 15.950  \\ \cdashline{2-6}
		& Average $\pm$ SD     & 15.014 $\pm$ 0.330 & 15.063 $\pm$ 0.380 & 16.190 $\pm$ 0.273 & 16.465 $\pm$ 0.296
		
	\end{tabular}
\end{table}

\begin{figure}[!bt]
	\centering
	\includegraphics[width=1.0\textwidth]{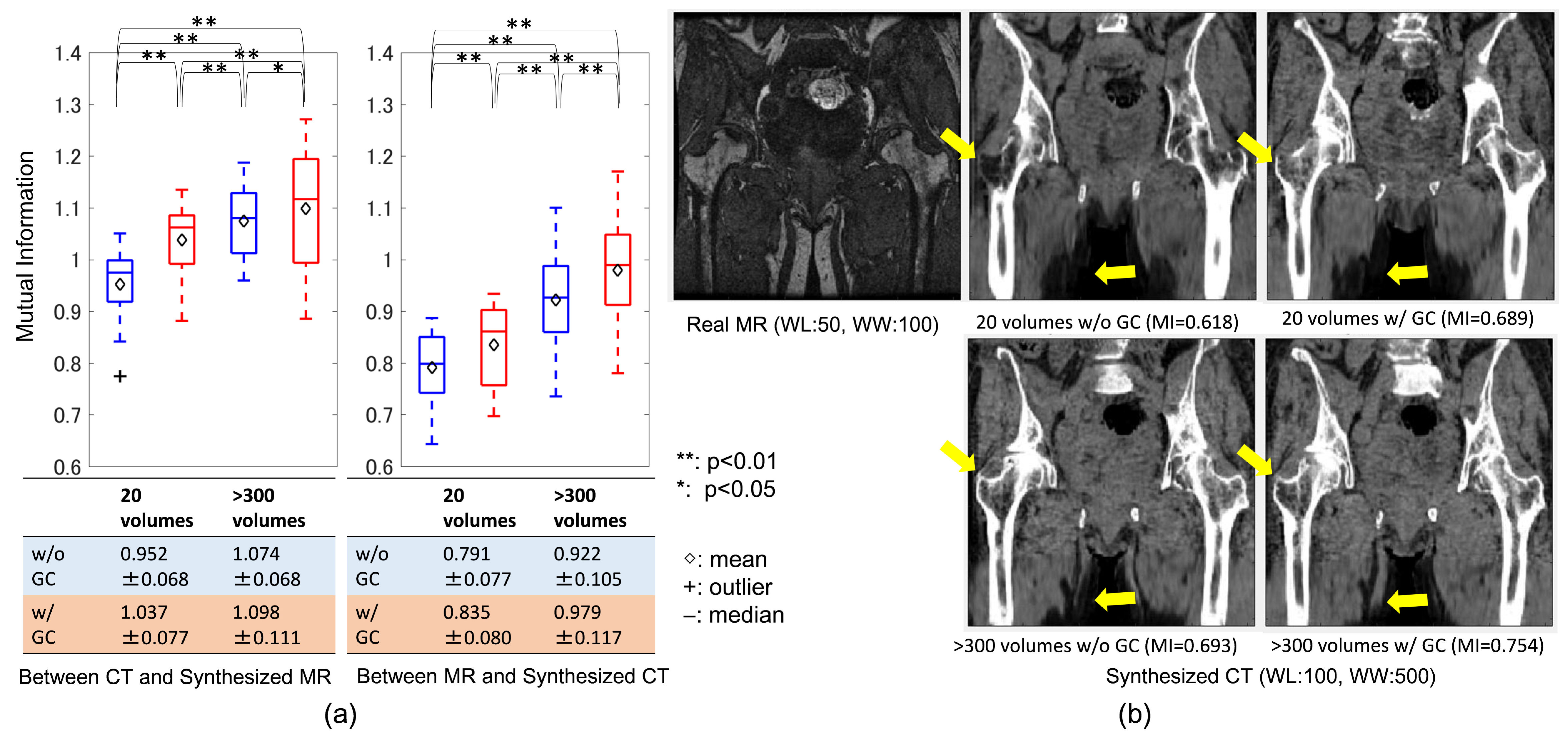}
	\caption{Evaluation of similarity between the real and synthesized volumes. (a) quantitative comparison of mutual information on different training data size with and without the gradient-consistency loss. (b) representative result of one patient. }
	\label{fig:eval_similarity}
\end{figure}

\begin{figure}[!bt]
	\centering
	\includegraphics[width=0.9\textwidth]{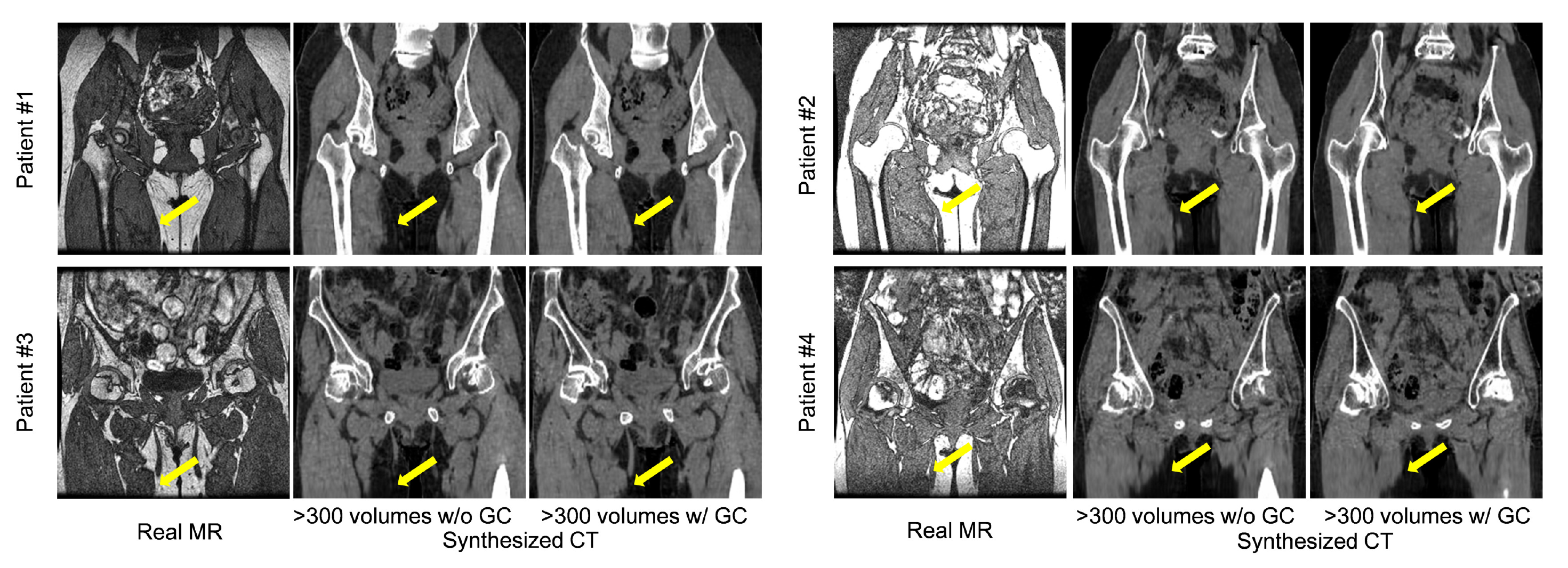}
	\caption{Representative results of translation from real MR to synthesized CT of four patients with and without the gradient consistency loss. As indicated by arrows, synthesized volumes with gradient consistency loss helped to preserve the shape near the adductor muscles. }
	\label{fig:vis_synthesis}
\end{figure}

In the second experiment, we tested with unpaired 10 MR and 20 CT volumes. 
Mutual information (MI) between synthesized CT and original MR was used for evaluation when the paired ground truth was not available.
The quantitative results are show in Fig.\ref{fig:eval_similarity}(a). The left side is the box and whisker plots of the mean of each slice of MI between real CT and synthesized MR (i.e., 20 data points in total). The right side is the mean of MI between real MR and synthesized CT (i.e., 10 data points in total). The result shows that the larger number of training data yielded statistically significant improvement ($p<0.01$) according to the paired $t$-test in MI. The GC loss also leads to an increase in MI between MR and synthesized CT ($p<0.01$). Fig.\ref{fig:eval_similarity}(b) and Fig.\ref{fig:vis_synthesis} show examples of the visualization of real MR and synthesized CT volumes. As indicated by arrows, we can see that synthesized volumes with GC loss preserved the shape near the femoral head and adductor muscles.

\subsection{Quantitative evaluation on segmentation}

To demonstrate the applicability of image synthesis in segmentation task, we evaluated the segmentation accuracy. Twenty labeled CT datasets were used to train the segmentation network. Then, we evaluated the segmentation accuracy with 10 MR volumes with manual segmentation labels of the gluteus medius and minimus muscles and femur.

We employed the 2D U-net proposed by Ronneberger et al. \cite{ronneberger2015u} as segmentation network, which is widely used in medical image analysis and demonstrated high performance with a limited number of labeled volumes. In MRI, muscle boundaries are clearer while bone boundaries are clearer in CT. To incorporate the advantage of both CT and MR, we modified the 2D U-net to  take the two-channel input of both CT and synthesized MR images.
We trained on 2D U-net using Adam \cite{kingma2014adam} for $1\times10^5$ iterations at learning rate of 0.0001. At the test phase, a pair of MR and synthesized CT was used as two-channel input.

\begin{figure}[!bt]
	\centering
	\includegraphics[width=0.85\textwidth]{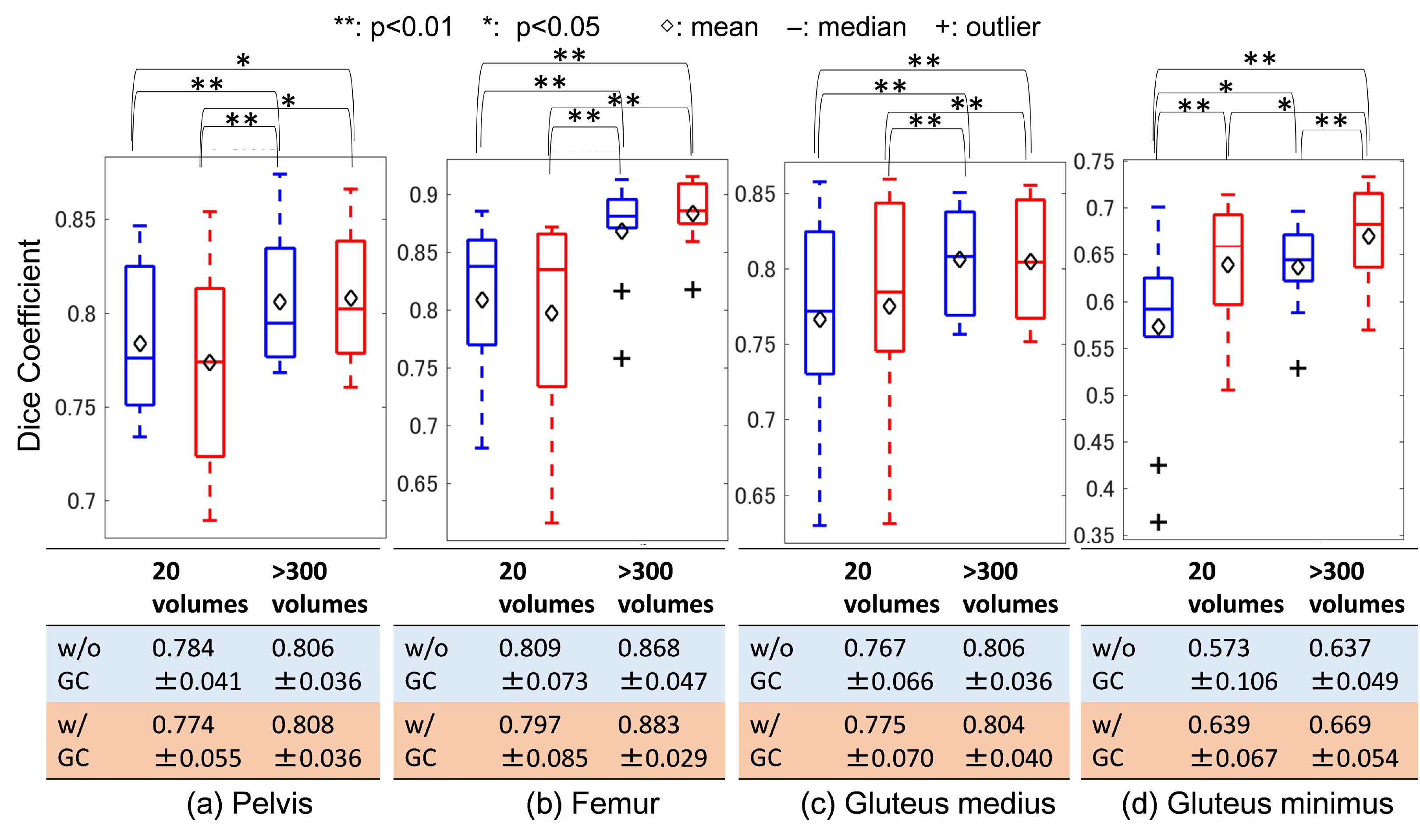}
	\caption{Evaluation of segmentation accuracy on different training data size in CycleGAN with and without the gradient-consistency loss. Segmentation of (a) pelvis, (b) femur, (c) gluteus medius and (d) gluteus minimus muscle in MR volumes were performed using MR-to-CT synthesis.}
	\label{fig:eval_seg}
\end{figure}

\begin{figure}[!bt]
	\centering
	\includegraphics[width=0.85\textwidth]{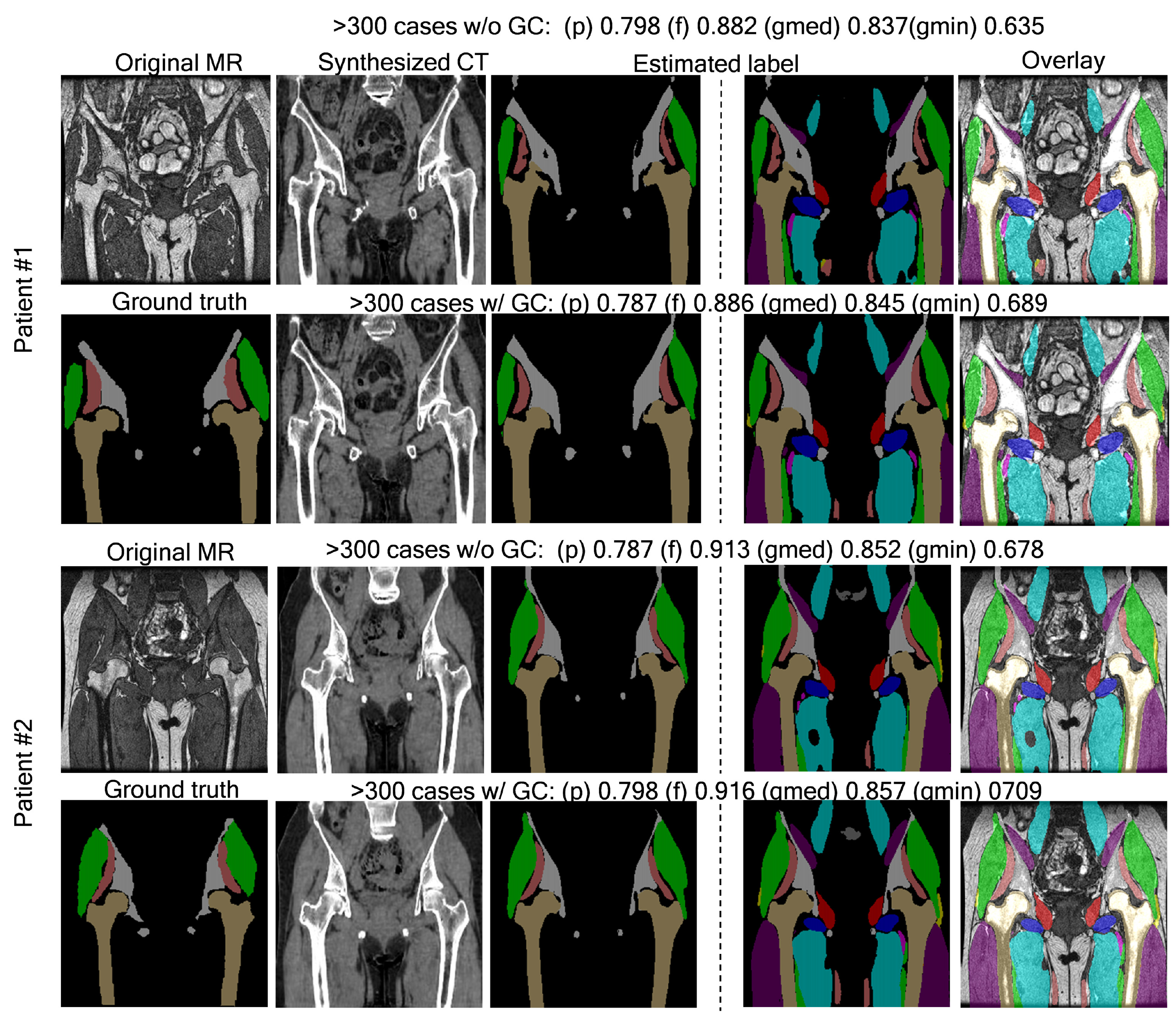}
	\caption{Representative results of segmentation from one patient. The ground truth label is consist of 4 musculoskeletal structures in MRI. Although we evaluated only on 4 structures because ground truth were not available for the other structures on MRI, all 22 estimated labels are shown for qualitative evaluation. In the right-most column,  all estimated labels are overlayed on the real MRI. p, f, gmed, gmin denote DICE of pelvis, femur, gluteus medius, and gluteus minimus, respectively.}
	\label{fig:vis_segment}
\end{figure}

The results with 4 musculoskeletal structures for 10 patients are shown in Fig.\ref{fig:eval_seg} (i.e., 10 data points in total on each plot). 
The result shows that the larger number of training data yielded statistically significant improvement in DICE on pelvis ($p<0.01$), femur ($p<0.01$), glutes medius  ($p<0.01$) and glutes minimus regions ($p<0.05$) of paired $t$-test. The GC loss also leads to an increase in DICE on the glutes minimus regions ($p<0.01$).
The average DICE coefficient in the cases trained with more than 300 cases and GC loss was 0.808$\pm$0.036 (pelvis), 0.883$\pm$0.029 (femur), 0.804$\pm$0.040 (gluteus medius) and 0.669$\pm$0.054 (gluteus minimus), respectively.
Fig.\ref{fig:vis_segment} shows example visualization of real MR, synthesized CT, and esimated label for one patient. 
The result with GC loss has smoother segmentation not only in the gluteus minimus but also near the adductor muscles. 

\section{Discussion and Conclusion} \label{sec:discussion section}

In this study, we proposed an image synthesis method which extended the CycleGAN approach by adding the GC loss to improve the accuracy at the boundaries. Specifically, the contributions of this paper are 1) introduction of GC loss in CycleGAN, and 2) quantitative and qualitative evaluation of the dependency of both image synthesis accuracy and segmentation accuracy on a large number of training data. 
One limitation in this study is that we excluded the patients with implants, while our target cohort (i.e., THA patients) sometime has implant on one side, for example, in case of the planning of secondary surgery.
As a comparison against a single modality training, we performed 5-fold cross validation of MR segmentation using 10 labeled MR volumes (i.e., trained with 8 MR volumes and tested on remaining 2 MR volumes) using U-net segmentation network. The DICE was 0.815$\pm$0.046 (pelvis), 0.921$\pm$0.023 (femur), 0.825$\pm$0.029 (gluteus medius) and 0.752$\pm$0.045 (gluteus minimus), respectively. 
We found the gap of accuracy between modality independent and dependent segmentation. A potential improvement of modality independent segmentation is to construct an end-to-end network that performs image synthesis and segmentation \cite{huo2017adversarial}.
Our future work also includes development of a method that effectively incorporates information in unlabeled CT and MR volumes to improve segmentation accuracy \cite{zhang2017deep}. 


\bibliography{refs}

\begin{thebibliography}{10}

\bibitem{cvitanic2004mri}
Cvitanic, O.,  et~al.:
\newblock {MRI} diagnosis of tears of the hip abductor tendons (gluteus medius
  and gluteus minimus).
\newblock American Journal of Roentgenology \textbf{182}(1) (2004)  137--143

\bibitem{torrado2016fast}
Torrado-Carvajal, A.,  et~al.:
\newblock Fast patch-based pseudo-{CT} synthesis from {T1}-weighted {MR} images
  for {PET/MR} attenuation correction in brain studies.
\newblock Journal of Nuclear Medicine \textbf{57}(1) (2016)  136--143

\bibitem{zhao2017whole}
Zhao, C.,  et~al.:
\newblock Whole brain segmentation and labeling from {CT} using synthetic {MR}
  images.
\newblock In: International Workshop on Machine Learning in Medical Imaging,
  Springer (2017)  291--298

\bibitem{kamnitsas2017unsupervised}
Kamnitsas, K.,  et~al.:
\newblock Unsupervised domain adaptation in brain lesion segmentation with
  adversarial networks.
\newblock In: International Conference on Information Processing in Medical
  Imaging, Springer (2017)  597--609

\bibitem{zhu2017unpaired}
Zhu, J.Y.,  et~al.:
\newblock Unpaired image-to-image translation using cycle-consistent
  adversarial networks.
\newblock In: Proceedings of the IEEE Conference on Computer Vision and Pattern
  Recognition. (2017)  2223--2232

\bibitem{wolterink2017deep}
Wolterink, J.M.,  et~al.:
\newblock Deep {MR} to {CT} synthesis using unpaired data.
\newblock In: International Workshop on Simulation and Synthesis in Medical
  Imaging, Springer (2017)  14--23

\bibitem{gilles2010musculoskeletal}
Gilles, B.,  et~al.:
\newblock Musculoskeletal {MRI} segmentation using multi-resolution simplex
  meshes with medial representations.
\newblock Medical image analysis \textbf{14}(3) (2010)  291--302

\bibitem{ranzini2017joint}
Ranzini, M.B.M.,  et~al.:
\newblock Joint multimodal segmentation of clinical {CT} and {MR} from hip
  arthroplasty patients.
\newblock In: International Workshop and Challenge on Computational Methods and
  Clinical Applications in Musculoskeletal Imaging, Springer (2017)  72--84

\bibitem{hamarneh2008simulation}
Hamarneh, G.,  et~al.:
\newblock Simulation of ground-truth validation data via physically-and
  statistically-based warps.
\newblock In: International Conference on Medical Image Computing and
  Computer-Assisted Intervention, Springer (2008)  459--467

\bibitem{tustison2010n4itk}
Tustison, N.J.,  et~al.:
\newblock {N4ITK}: improved {N}3 bias correction.
\newblock IEEE transactions on medical imaging \textbf{29}(6) (2010)
  1310--1320

\bibitem{penney1998comparison}
Penney, G.P.,  et~al.:
\newblock A comparison of similarity measures for use in 2-{D}-3-{D} medical
  image registration.
\newblock IEEE transactions on medical imaging \textbf{17}(4) (1998)  586--595

\bibitem{johnson2016perceptual}
Johnson, J.,  et~al.:
\newblock Perceptual losses for real-time style transfer and super-resolution.
\newblock In: European Conference on Computer Vision, Springer (2016)  694--711

\bibitem{isola2017image}
Isola, P.,  et~al.:
\newblock Image-to-image translation with conditional adversarial networks.
\newblock arXiv preprint (2017)

\bibitem{mao2016multi}
Mao, X.,  et~al.:
\newblock Multi-class generative adversarial networks with the {L2} loss
  function.
\newblock CoRR, abs/1611.04076 \textbf{2} (2016)

\bibitem{kingma2014adam}
Kingma, D.P.,  et~al.:
\newblock Adam: A method for stochastic optimization.
\newblock arXiv preprint arXiv:1412.6980 (2014)

\bibitem{ronneberger2015u}
Ronneberger, O.,  et~al.:
\newblock U-net: Convolutional networks for biomedical image segmentation.
\newblock In: International Conference on Medical image computing and
  computer-assisted intervention, Springer (2015)  234--241

\bibitem{huo2017adversarial}
Huo, Y.,  et~al.:
\newblock Adversarial synthesis learning enables segmentation without target
  modality ground truth.
\newblock arXiv preprint arXiv:1712.07695 (2017)

\bibitem{zhang2017deep}
Zhang, Y.,  et~al.:
\newblock Deep adversarial networks for biomedical image segmentation utilizing
  unannotated images.
\newblock In: International Conference on Medical Image Computing and
  Computer-Assisted Intervention, Springer (2017)  408--416

\end{thebibliography}
\bibliographystyle{splncs_03} 
	
\end{document}